 \documentclass[pmlr,twocolumn,10pt]{jmlr} 
 \colorlet{promptmint}{green!30}
\usepackage{booktabs}
\usepackage{makecell}
\usepackage{longtable}
\usepackage{cuted}  
\usepackage{enumitem}
\usepackage[most]{tcolorbox}
\usepackage{xcolor}
\usepackage{lmodern}
\usepackage{array}
 
\usepackage{ragged2e} 
\newcolumntype{L}[1]{>{\RaggedRight\arraybackslash}p{#1}}
\usepackage{siunitx}
\usepackage{booktabs}
\usepackage{amssymb,pifont}   
\usepackage{pgfplots}
\usepackage{float}
\pgfplotsset{compat=1.18}
\usepackage{tikz}
\usetikzlibrary{positioning,arrows.meta,shapes.geometric,shapes.misc,fit,backgrounds}

\usepackage{microtype}
\usepackage{caption} 
\usepackage{xcolor,pifont}
\newcommand{\cmark}{\textcolor{green!60!black}{\ding{51}}} 
\newcommand{\xmark}{\textcolor{red!70!black}{\ding{55}}}   
\usepackage{multirow} 

\usepackage{siunitx}
 \usepackage{pifont}
\usepackage[switch]{lineno}
\nolinenumbers
\usepackage{afterpage}
\usepackage{etoolbox}
\makeatletter
\AtBeginEnvironment{thebibliography}{%
  \begingroup
  \raggedbottom
  \setlength{\bibsep}{0pt plus .2ex minus .2ex}
  \setlength{\itemsep}{0pt plus .2ex minus .2ex}
  \setlength{\parskip}{0pt}
}
\AtEndEnvironment{thebibliography}{\endgroup}
\makeatother

\usepackage{xurl}
\usepackage{placeins}
\usepackage[utf8]{inputenc}

\theorembodyfont{\upshape}
\theoremheaderfont{\scshape}
\theorempostheader{:}
\theoremsep{\newline}

\usepackage{makecell}
\jmlrvolume{297}
\jmlrworkshop{Machine Learning for Health (ML4H) 2025} 

 \title[MIMIC-SR-ICD11: A Dataset for Narrative-Based Diagnosis]{MIMIC-SR-ICD11: A Dataset for Narrative-Based Diagnosis}

\author{%
\Name{Yuexin Wu} \Email{ywu10@memphis.edu}\\
\addr University of Memphis, Unite State
\AND 
\Name{Shiqi Wang} \Email{20251110858@stu.gzucm.edu.cn}\\
\addr The Second Clinical College of Guangzhou University of Chinese Medicine, China
\AND
\Name{Vasile Rus} \Email{vrus@memphis.edu}\\
\addr University of Memphis, Unite State
}

\begin{document}
\nolinenumbers
\maketitle

\begin{abstract}
Disease diagnosis has become a central pillar of modern healthcare, enabling early detection and timely intervention for acute conditions while guiding lifestyle adjustments and medication regimens to prevent or slow chronic disease.  Self-reports preserve clinically salient signals that templated electronic health record (EHR) documentation often attenuates or omits, especially subtle but consequential details. To operationalize this shift, we introduce MIMIC\textendash SR\textendash ICD11, a large English diagnostic dataset built from EHR discharge notes and natively aligned to WHO ICD-11 terminology. We further present LL-Rank, a likelihood based re-ranking framework that compute a length normalized joint likelihood of each label under the clinical report context and subtracting the corresponding report free prior likelihood of that label. On seven model backbones, LL\textendash Rank consistently outperforms a strong generation $+$ mapping baseline (GenMap). Ablation experiments reveal that LL–Rank’s improvement mainly arises from its PMI-based scoring, which isolates semantic compatibility from label frequency bias. 
\end{abstract}


\paragraph*{Data and Code Availability}
Code is available at \url{https://github.com/woqingdoua/MIMIC-SR-ICD11}. Data use de-identified MIMIC-IV/MIMIC-IV-Note under PhysioNet credentialed access; our generated self-reports are released for research-only use.

\paragraph*{Institutional Review Board (IRB)}
This study did not involve human subjects or identifiable private information and therefore did not require IRB approval.

\section{Introduction}
Disease diagnosis has become a central pillar of modern healthcare, enabling early detection and timely intervention for acute conditions, while also guiding lifestyle adjustments and medication regimens to prevent or slow chronic diseases. It is particularly valuable in resource-limited environments and helps individuals without medical expertise avoid a long search for the right provider.

More recently, large language models (LLMs) have demonstrated strong performance on clinical question–answering benchmarks~\cite{Singhal2023Large, Singhal2025Toward}. These models are typically fine-tuned on exam-style question–answer datasets designed for medical students~\cite{jin2020disease,pal2022medmcqa,jin2019pubmedqa}, but their training regime does not directly translate to real-world diagnostic workflows, since exam-style benchmarks present well-defined questions with fixed answer options, whereas clinical diagnosis requires interpreting ambiguous, multi-symptom narratives. Datasets for automatic diagnosis such as DX~\cite{xu2019end} and DDXPlus~\cite{fansi2022ddxplus} rely on categorical symptom indicators. This representation obscures important clinical detail—for example, reducing “severe, intermittent chest pain radiating to the left arm” to a present/absent flag loses information about intensity and distribution—and, because these collections are built for fixed-label classification, models trained on them cannot readily incorporate new symptoms or expand beyond the original disease set. Su et al.~\cite{su2024enabling} take a step toward free-text inputs with patient-authored symptom descriptions, but their dataset is confined to Chinese and leaves English self-reports unexplored.

Taken together, these limitations motivate a shift toward inputs that mirror how patients present symptoms at first contact: patient-authored self-reports. Self-reports can be collected online before a visit and are well suited to telemedicine, longitudinal follow-up, and population screening. They are especially effective for early triage and first-contact routing, and they generalize across front-end applications such as online triage tools, navigation assistants, and conversational agents. Critically, self-reports preserve clinically salient detail that templated EHR documentation often attenuates or omits (e.g. the time course of illness, subjective experience (pain intensity, triggers and relievers), subtle yet important co-occurring symptoms). Training and evaluating on self-reports therefore better matches deployment conditions and captures information that categorical checklists or templated notes tend to suppress.

To operationalize this shift, we introduce an English-language dataset, MIMIC-SR-ICD11, which converts EHR discharge notes into first-person patient self-reports and standardizes diagnoses using WHO ICD-11 terminology. Building on the advantages of a unified label space, we propose LL-Rank, a likelihood-based re-ranking method that combines the conditional likelihood of each ICD-11 label given the report with an explicit corpus-derived prior. By balancing textual evidence with label priors, LL-Rank yields better-calibrated rankings and consistently improves performance, with particularly strong gains on short-token labels. 

\section{Data Construction}
We built the dataset on the latest MIMIC-IV~\cite{Johnson2024MIMICIV} and MIMIC-IV-Note~\cite{Johnson2022MIMICIVNote} releases. The construction pipeline consists of two steps, with the workflow shown in Fig.~\ref{fig:data-pipeline}. First, 
we map ICD-9 and ICD-10 diagnoses from MIMIC-IV to ICD-11 terminology. Second, we retrieve free-text symptom descriptions from the MIMIC-IV-Note EMR notes and rewrite them as first-person patient self-reports.

\subsection{Diagnosis Mapping}
We derive diagnostic labels from MIMIC-IV and retain only the primary diagnosis. Each record carries an ICD code and a version flag. For encounters coded in ICD-9, we first convert to ICD-10 using the U.S. Centers for Medicare \& Medicaid Services (CMS) official FY2018 General Equivalence Mappings (GEMs)\footnote{\url{https://www.cms.gov/medicare/coding-billing/icd-10-codes}}. In this step, we keep only entries that are both “exact” and “mappable” in the GEMs metadata. We then enforce a one-to-one constraint at the code level.  ICD-9 codes that map to multiple ICD-10 candidates are withheld for manual review rather than automatically converted.

Next, for all rows with an ICD-10 code (either originally ICD-10 or obtained from the previous step), we map to ICD-11 using the World Health Organization’s official mapping tables provided via the ICD-11 Browser \footnote{\url{https://icd.who.int/browse/2025-01/mms/en}}. We again retain only one-to-one correspondences to preserve semantic precision. After automatic mapping, approximately 10\% of diagnoses are fully resolved.  The remainder are curated by medically trained annotators who select the most faithful ICD-11 entity by consulting the source descriptor and validating candidates in the ICD-11 browser. During curation we exclude non-disease concepts (e.g., symptom/sign–only entries, aftercare, external causes) and overly broad or ambiguous categories to ensure labels represent clinically actionable diseases.

\subsection{Patient's self-report generation}

The MIMIC-IV-Note dataset provides de-identified free-text hospital records for each patient, which typically include a mix of symptom descriptions, examination results, medical history, and social background information. To derive patient-style narratives suitable for large language model reasoning, we used ChatGPT\footnote{\url{https://openai.com/api/}} (gpt-4o-mini) to convert these clinical notes into first-person self-reports. 
Before large-scale generation, our medical students curated a small development set of 10 notes and prompt-tuned the instruction (i.e., iteratively refined the wording and few-shot exemplars) so that the model reliably transforms EHR notes into patient-style narratives. During conversion, the prompt explicitly instructs the model to filter out clinician-generated content (physical examination findings, diagnostic test results, and professional assessments) and to retain only subjective symptom descriptions as recounted by the patient. The resulting self-reports are written in natural language using complete sentences from the patient’s perspective, which better aligns the input with real-world patient narratives and facilitates downstream disease diagnosis.


\section{Method}

\subsection{Setup and Notation}
Let $x$ denote a patient self–report and let $\mathcal{C}=\{c_j\}_{j=1}^{M}$ be the fixed candidate set of ICD–11.
We fine–tune LLMs with supervised fine-tuning (SFT) and LoRA adapters, then keep the adapted model $\theta$ for inference.
Given a patient's report,
we wish to rank $\mathcal{C}$ and evaluate top–$k$ predictions.

\subsection{Method 1: Greedy generation + label mapping (GenMap)}
We first prompt the model $\theta$ to generate a short diagnostic phrase using deterministic greedy decoding. The resulting text is then mapped to the closest disease name from a fixed candidate list. To measure closeness, we compute token-level overlap between the generated phrase and each candidate label after tokenization. Each shared token contributes inversely to its overall frequency in the label corpus, so rarer and thus more informative tokens receive higher weight.

Candidates are ranked by a two-part key: (1) the number of overlapping tokens, and (2) the weighted rarity score used to break ties. This procedure favors candidates sharing distinctive terms with the generated text while de-emphasizing common medical words such as “disease” or “unspecified.” Sorting all candidates by this key yields a complete ranking, denoted GenMap. We report standard retrieval metrics including Hit@ k and Macro-F1$@$k based on the top-k predictions.

\subsection{Method 2: PMI-style Scoring (LL-Rank)}
Rather than decoding a free string, we directly score each diagnosis candidate $c\in\mathcal{C}$ with a Pointwise Mutual Information (PMI) style criterion. 
PMI measures how much knowing $x$ changes the likelihood of $c$. Using PMI rather than $\log p(c\mid x)$ alone discounts labels that are a priori frequent, thereby focusing the score on compatibility with the input rather than corpus popularity.

Given a candidate label \(c\) with tokenization \(\tau(c)=y_{1:T(c)}\) and a fixed textual prompt \texttt{prefix}, we score it by the conditional per-token negative log-likelihood $\mathcal{L}_{\mathrm{cond}}(x,c)$.
By the autoregressive factorization, the label likelihood decomposes into a product over its tokens.  
We therefore aggregate token log-likelihoods and divide by \(T(c)\) to obtain a per-token (length-normalized) score, ensuring labels with more tokens are not unfairly penalized.

To mitigate prior bias (i.e., the tendency to overestimate very frequent labels), we also compute a report-free NLL using only the prefix context:
\begin{align*}
\mathcal{L}_{\mathrm{prior}}(c)
  &= \frac{1}{T(c)} \sum_{t=1}^{T(c)}
     \Bigl[-\log p_{\theta}\!\big(y_t \,\big|\, \text{\texttt{prefix}},\, y_{<t}\big)\Bigr].
\label{eq:lprior}
\end{align*}
This term captures how intrinsically common a label is under the model (with identical decoding setup), on the same per-token scale. 

Finally, we combine the two losses into a PMI-style score by subtracting the report-free (prior) term from the report-conditioned term, with a nonnegative weight ($\alpha$):
\begin{align*} 
S(x,c) &= -\,\mathcal{L}_{\mathrm{cond}}(x,c) \;+\; \alpha\,\mathcal{L}_{\mathrm{prior}}(c), \alpha \ge 0
\end{align*}
Candidates are then sorted in descending order of (S(x,c)), and the top-(k) labels are returned as predictions. This construction explicitly removes the model’s unconditional tendency to emit high-frequency labels while keeping the two likelihoods comparable because they share the same prompt, tokenization, and length normalization. In all experiments we set ($\alpha=1$), which offers a good trade-off: ($\alpha=0$) collapses to pure conditional likelihood (and thus reintroduces frequency bias), whereas ($\alpha>1$) can over-penalize labels that are common but still correct.

\begin{table*}[t]
\centering
\caption{Comparison of two scoring methods. For each model we show M1 (GenMap), M2 (LL-Rank), and the relative improvement $\Delta = \frac{\text{M2} - \text{M1}}{\text{M1}}$. $\bar{\Delta}$ is the mean percentage improvement across all models.}
\small
\begin{tabular}{llcccccc}
\toprule
\multirow{2}{*}{Model} & \multirow{2}{*}{Scoring} &
\multicolumn{2}{c}{Top-3} & \multicolumn{2}{c}{Top-5} & \multicolumn{2}{c}{Top-10} \\
\cmidrule(lr){3-4}\cmidrule(lr){5-6}\cmidrule(lr){7-8}
 &  & Hit@3 & Macro-F1 & Hit@5 & Macro-F1 & Hit@10 & Macro-F1 \\
\midrule
\multirow{3}{*}{MedAlpaca (7B)}
 & M1: GenMap   & 29.52 & 16.00 & 34.02 & 14.67 & 40.42 & 11.91 \\
 & M2: LL-Rank  & 67.99 & 35.86 & 76.92 & 31.02 & 84.53 & 23.81 \\
 & $\Delta$     & 130.3\% & 124.1\% & 126.1\% & 111.5\% & 109.1\% &  99.9\% \\
\midrule
\multirow{3}{*}{MedLLaMA (8B)}
 & M1: GenMap   & 26.99 & 14.18 & 32.74 & 14.28 & 43.61 & 13.35 \\
 & M2: LL-Rank  & 49.89 & 28.43 & 61.30 & 24.45 & 75.68 & 16.48 \\
 & $\Delta$     &  84.8\% & 100.5\% &  87.2\% &  71.2\% &  73.5\% &  23.4\% \\
\midrule
\multirow{3}{*}{MedGEMMA (3B)}
 & M1: GenMap   & 46.25 & 26.11 & 49.13 & 22.74 & 53.68 & 16.31 \\
 & M2: LL-Rank  & 42.57 & 26.74 & 52.13 & 23.88 & 65.39 & 20.02 \\
 & $\Delta$     &  -8.0\% &   2.4\% &   6.1\% &   5.0\% &  21.8\% &  22.7\% \\
\midrule
\multirow{3}{*}{AlphaMed (3B)}
 & M1: GenMap   & 10.02 &  3.85 & 17.29 &  4.57 & 25.17 &  4.11 \\
 & M2: LL-Rank  & 20.27 & 17.75 & 32.05 & 24.08 & 58.67 & 27.46 \\
 & $\Delta$     & 102.3\% & 361.0\% &  85.4\% & 426.9\% & 133.1\% & 568.1\% \\
\midrule
\multirow{3}{*}{AlphaMed (7B)}
 & M1: GenMap   & 30.32 & 15.92 & 36.42 & 15.66 & 44.72 & 12.92 \\
 & M2: LL-Rank  & 53.41 & 32.56 & 65.67 & 31.04 & 80.08 & 24.27 \\
 & $\Delta$     &  76.2\% & 104.5\% &  80.3\% &  98.2\% &  79.1\% &  87.8\% \\
\midrule
\multirow{3}{*}{MedFound (7B)}
 & M1: GenMap   & 20.45 &  9.24 & 24.01 &  8.39 & 29.61 &  7.07 \\
 & M2: LL-Rank  & 32.84 & 20.15 & 41.86 & 20.92 & 56.75 & 20.31 \\
 & $\Delta$     &  60.6\% & 118.1\% &  74.3\% & 149.3\% &  91.7\% & 187.3\% \\
\midrule
\multirow{3}{*}{MedFound (8B)}
 & M1: GenMap   & 18.94 & 11.27 & 22.69 & 11.05 & 30.87 & 10.53 \\
 & M2: LL-Rank  & 40.57 & 28.81 & 54.63 & 29.20 & 74.56 & 21.83 \\
 & $\Delta$     & 114.2\% & 155.6\% & 140.8\% & 164.3\% & 141.5\% & 107.3\% \\
\midrule
$\bar{\Delta}$ && 80.1\% & 138.0\% & 85.7\% & 146.6\% & 92.8\% & 156.7\% \\
\bottomrule
\end{tabular}
\label{tab:m1vsm2}
\end{table*}

\section{Results}

\subsection{GenMap (M1) vs.\ LL-Rank (M2)}
We compare two prediction methods (GenMap and LL-Rank), and present the results in Table~\ref{tab:m1vsm2}. Across seven backbones, the LL-Rank (M2) method consistently outperforms the GenMap (M1) baseline on nearly all metrics. Averaged over models, M2 increases Hit@{3,5,10} by 80.1\%, 85.7\%, and 92.8\%, and boosts Macro–F1@{3,5,10} by 138.0\%, 146.6\%, and 156.7\%, respectively. The gains are especially pronounced for Macro–F1, indicating that M2 strengthens performance on underrepresented labels rather than merely amplifying head classes. Improvements also grow from Top–3 through Top–10, suggesting that M2 not only raises the likelihood of retrieving a correct label early, but also produces a higher-quality ranked list overall.

\begin{figure}
  \centering
  \includegraphics[width=0.98\linewidth]{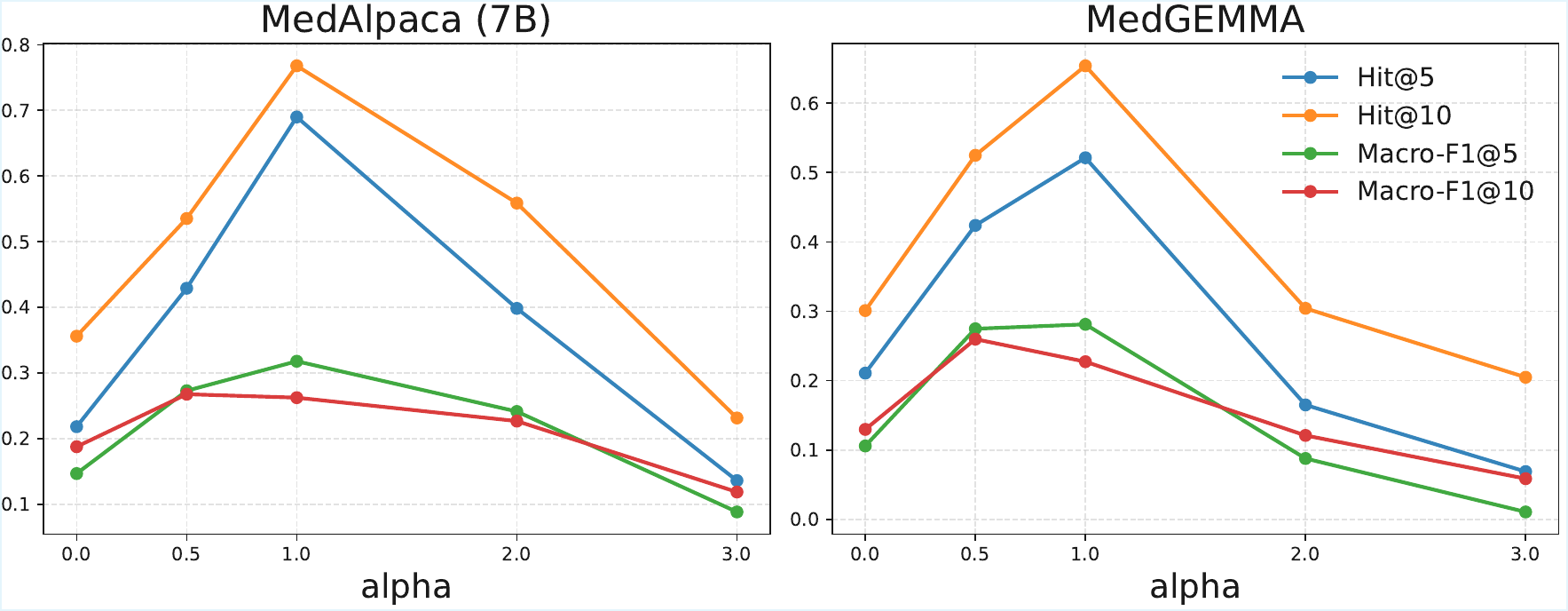}
  \caption{Effect of the PMI coefficient ($\alpha$) on LL-Rank scoring.}
  \label{fig:alpha_sweep}
\end{figure}

\subsection{Effect of the PMI Coefficient }
Figure~\ref{fig:alpha_sweep} examines how the LL-Rank method behaves as the PMI prior weight $\alpha$ varies. LL-Rank’s performance is highly sensitive to the prior weight $\alpha$. Across both backbones (MedAlpaca--7B and MedGEMMA--3B) performance peaks near $\alpha\!\approx\!1$: when $\alpha$ is smaller, the prior under-corrects head-class bias and conditional likelihood dominates, leaving rare labels undervalued; when $\alpha$ is larger, the prior is over-applied, penalizing frequent labels too strongly and degrading the ranked list. The region around $\alpha\!\approx\!1$ consistently yields the best Hit and Macro F1. Compared with GenMap’s token-rarity heuristic, injecting an explicit prior via LL--Rank is markedly more effective for the long tail. Relative to no prior (\(\alpha=0\)), setting \(\alpha=1\) roughly doubles performance on hit rate and macro F1.

\section{Conclusion}
We release MIMIC\textendash SR\textendash ICD11, a large English diagnostic dataset built from patient-authored self-reports and natively aligned with WHO ICD\textendash 11. On this resource,  we instantiate LL–Rank, a likelihood-based re-ranking method that combines the conditional per-token likelihood of each label given the report with an explicit corpus-derived prior, counteracting head-class bias. Across seven backbones, LL–Rank consistently outperforms GenMap. 

\clearpage

\bibliography{jmlr-sample}

\newpage
\appendix


\begin{strip}
  \centering
  \includegraphics[width=\textwidth]{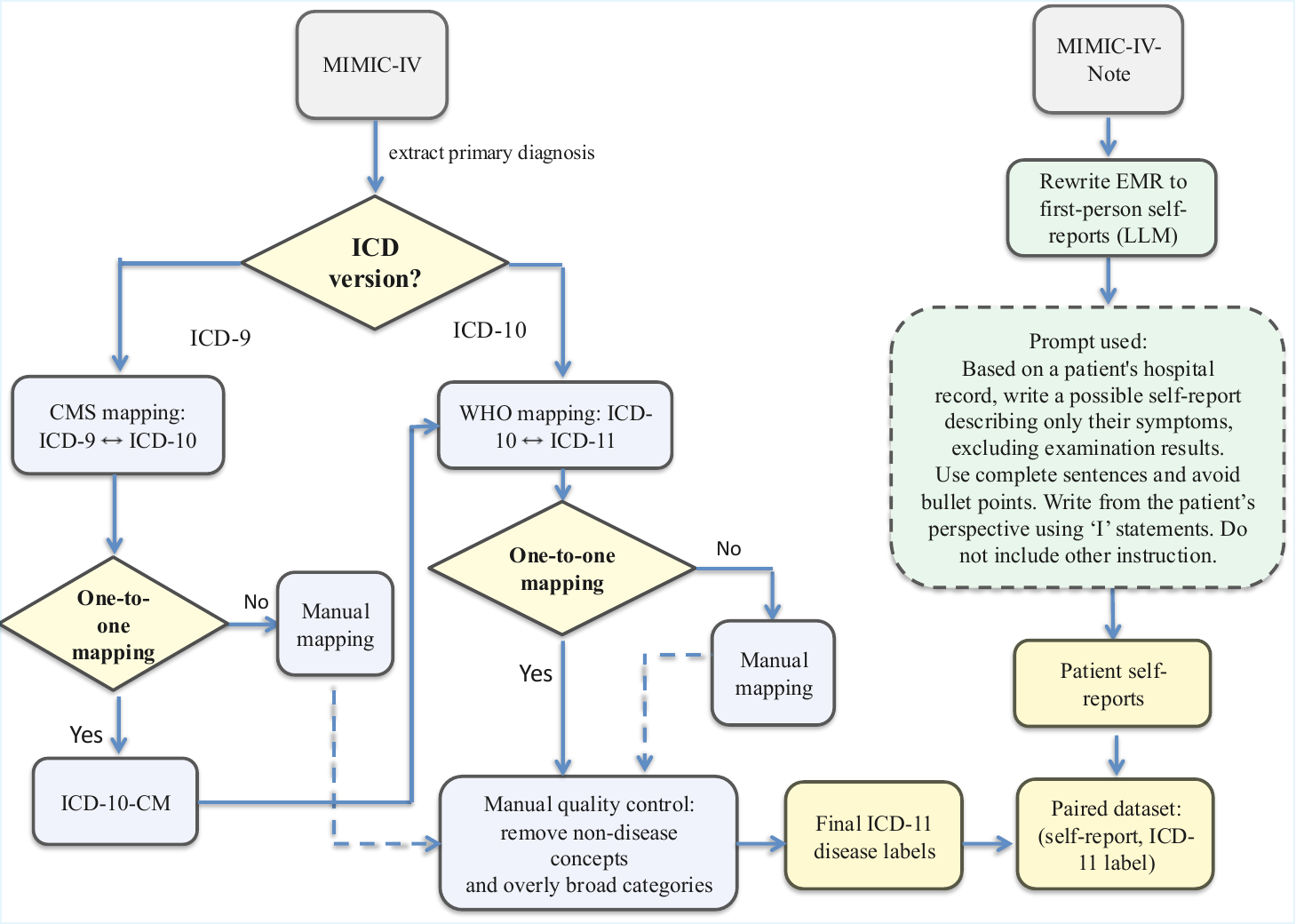}
  \captionof{figure}{Data construction pipeline. Left branch: extract primary diagnoses from MIMIC-IV and map ICD-9 to ICD-10 then ICD-10 to ICD-11, with one-to-one filtering and manual quality control. Right branch: rewrite MIMIC-IV-Note into first-person self-reports using ChatGPT. The dashed box shows the exact prompt we used in practice. Outputs: final ICD-11 labels and paired records.}
  \label{fig:data-pipeline}
\end{strip}

\section{Dataset Construction and Preprocessing}
Figure~\ref{fig:data-pipeline}. 

\begin{table*}[t]
\centering
\small
\caption{Comparison of diagnosis datasets.Columns:
Dis.\# = number of disease categories; Real = real-world provenance; Gen. = generalist coverage; SR = patient self-report; Free-text = unstructured input; Ont = ontology alignment (e.g., ICD-11); Lang = language.}
\begin{tabular}{l c c c c c c c c}
\hline
\textbf{Dataset / Benchmark} & \textbf{Size} & \textbf{Dis.\#} & \textbf{Real} & \textbf{Gen.} & \textbf{SR} & \textbf{Free-text} & \textbf{Ont} & \textbf{Lang} \\
\hline
Muzhi~\cite{wei2018task}                 & 710          & 4   & \xmark & \xmark & \cmark & \cmark & \xmark         & ZH \\
DX~\cite{xu2019end} & 527 & 5 & \cmark & \xmark & \cmark & \cmark & \xmark & ZH \\
DDXPlus~\cite{fansi2022ddxplus}          & 1{,}300{,}000 & 49  & \xmark & \xmark & \xmark & \xmark & \xmark         & EN \\
NLICE~\cite{al2023nlice}                 & 1{,}000{,}000 & 55  & \xmark & \xmark & \xmark & \xmark & \xmark         & EN \\
RareBench~\cite{chen2024rarebench}       & 2{,}185      & 421 & \cmark & \xmark & \xmark & \cmark & \cmark         & ZH/EN \\
MSDiagnosis~\cite{hou2024msdiagnosis}    & 2{,}225      & --- & \cmark & \cmark & \xmark & \cmark & \xmark         & ZH \\
Haodf~\cite{su2024enabling}              & 29{,}326     & 190 & \cmark & \cmark & \cmark & \cmark & \xmark         & ZH \\
SDBench~\cite{nori2025sequential}        & 304          & --- & \cmark & \cmark & \xmark & \cmark & \xmark         & EN \\
Open-XDDx~\cite{zhou2025explainable} & 570   & --- & \xmark & \cmark & \xmark & \cmark & \xmark         & EN \\
CMEMR~\cite{jia2025medikal}              & 10{,}450     & --- & \cmark & \cmark & \xmark & \cmark & \xmark         & ZH \\
MIMIC\textendash SR\textendash ICD11 (Ours) & 119{,}178  & 118 & \cmark & \cmark & \cmark & \cmark & \cmark         & EN \\
\hline
\end{tabular}
\label{tab:dx_benchmarks_realworld}
\end{table*}
\section{Related Work}
 In this section we review recent diagnosis-oriented datasets. We organize the landscape into two streams: (i) symptom-inquiry policy learning, which focuses on interactive, slot-based symptom collection, and (ii) full-text diagnostic inference, which infers diseases from unstructured narratives. Table~\ref{tab:dx_benchmarks_realworld} summarizes key properties (e.g. scale, disease coverage, real world provenance, support for patient self-reports (SR), free-text availability, ontology alignment, language) and contrasts them with our proposed MIMIC–SR–ICD11.
 
\paragraph{Symptom-inquiry policy learning}
Symptom-inquiry policy learning views the clinical interview as sequential evidence gathering: the system chooses which symptom to ask about next until it has enough information to issue a diagnosis. Representative datasets for this scenario include Muzhi~\cite{wei2018task}, DX, DDXPlus~\cite{fansi2022ddxplus}, NLICE~\cite{al2023nlice}  , RareBench~\cite{chen2024rarebench}, SDBench~\cite{nori2025sequential}, and Open-XDDx~\cite{zhou2025explainable}. Muzhi introduces small task-oriented diagnostic dialogues centered on follow-up symptom questioning. DX builds on this by moving from curated toy dialogues to real-world online consultations with free-text narratives. RareBench shifts the focus to rare diseases and standardizes labels via clinical ontologies, enabling standardized evaluation. SDBench then changes the mode from static dialogue to a step-by-step case: at each step the system can ask more history or order a test, and every test has a price tag—the aim is to reach the correct diagnosis while spending as little as possible. Open-XDDx keeps the narrative format and adds expert rationales, allowing assessment of both accuracy and the quality of explanations. Because these practice-derived resources are small (hundreds to a few thousand cases), they are best suited for prompting, few-shot evaluation, or lightweight instruction/preference tuning—often with synthetic augmentation—rather than stable supervised fine-tuning of billion-parameter LLMs. To address scale, synthetic cohorts such as DDXPlus and NLICE expand to hundreds of thousands or millions of encounters and enrich symptoms from binary indicators to multi-dimensional descriptors (e.g., location, intensity, duration), improving controllability and reproducibility while remaining compatible with sequential decision formulations; however, by omitting free-text self-reports they sacrifice realism, which can hinder direct transfer to narrative-driven diagnostic workflows.

\paragraph{Full-text Diagnostic Inference}
Unlike resources built for symptom-inquiry policy learning, full-text diagnostic inference (FTDI) evaluates models on complete clinical narratives: the system reads an entire note or consultation and then predicts the diagnosis. Datasets in this setting include MSDiagnosis~\cite{hou2024msdiagnosis}, Haodf~\cite{su2024enabling}, and CMEMR~\cite{jia2025medikal}. MSDiagnosis and CMEMR are Chinese EMR corpora for document-level diagnosis. In both, gold diagnoses are free text. Compared with MSDiagnosis (2.2k cases across 12 departments), CMEMR scales to 10.5k cases with broader specialty coverage and richer, longer documents—including admission notes, hospital course, and lab/imaging summaries—making it well suited for long-form diagnostic inference and information extraction. In contrast to these clinician-authored EMRs, Haodf is patient-side: it collects real online consultations with free-text self-reports and physician replies, covering many common conditions. It is valuable for generalist triage/diagnosis from the patient’s perspective. However, labels are not natively ontology aligned and the corpus is Chinese only, which complicates cross-dataset comparison and large-scale finetuning.


\begin{table*}
\centering
\small
\caption{Summary of baseline models.}
\resizebox{\linewidth}{!}{%
\begin{tabular}{l c c c c c}
\hline
\textbf{Model(s)} & \textbf{Size} & \textbf{Method} & \textbf{Objective} & \textbf{Data Source} & \textbf{Backbone} \\
\hline
MedAlpaca        & 7B        & SFT               & QA / dialogue              & med education                 & LLaMA v1 \\
MMedLLaMA        & 8B        & DAPT + SFT        & MCQ (with rationales)      & med knowledge, med education  & Llama 3 \\
MedGEMMA         & 3B        & VLP + SFT + RL    & VLM QA / report generation & med education                 & Gemma-3 \\
AlphaMed         & 3B / 7B   & RL (GRPO)         & MCQ (boxed final)          & med education                 & Qwen2 / Qwen2.5 \\
MedFound         & 7B / 8B   & DAPT + CoT + DPO  & Diagnostic reasoning       & clinical notes, med reference & BLOOM / Llama 3 \\
\hline
\end{tabular}}
\label{tab:baseline_models_resize}
\end{table*}

In summary, datasets for symptom-inquiry policy learning are mostly small, practice-derived free-text corpora. By contrast, DDXPlus and NLICE scale to hundreds of thousands or millions of cases but drop free text, limiting realism. Existing full-text diagnostic inference (FTDI) corpora are predominantly Chinese (e.g., MSDiagnosis, CMEMR, Haodf). Most resources above also lack native ontology alignment. Their labels are free-named and mapped to a standard vocabulary only at evaluation time. Compared with these datasets, our proposed MIMIC-SR-ICD11 is, to our knowledge, the first large-scale, English, patient self-report dataset for FTDI with native ICD-11 standardization. This design reduces linguistic ambiguity, removes fragile post-hoc mapping during evaluation, and offers generalist coverage (118 diagnoses) at benchmark scale (~119k reports). We release fixed splits and an evaluation script to enable reproducible training and fair comparison.

\section{Experiments}

\begin{figure}
  \centering
  \includegraphics[width=0.98\linewidth]{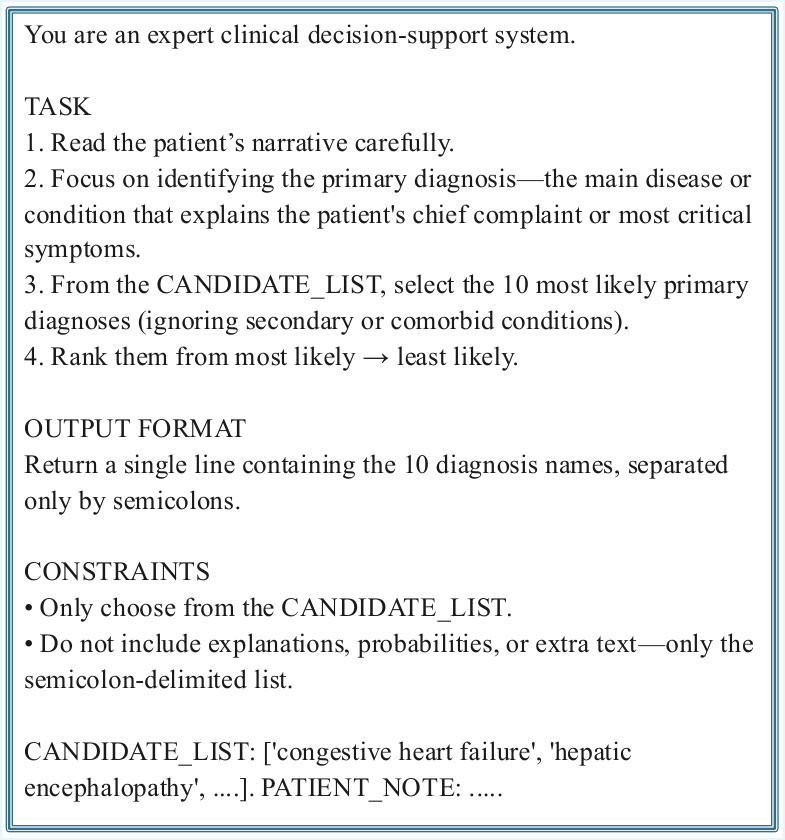}
   \caption{Illustration of the standardized prompt format (instruction, candidate list, and patient self-report) provided to the general LLMs.}
  \label{fig:prompt-workflow}
\end{figure}

\subsection{Baseline}

\subsubsection{Medical Domain LLMs}

MedAlpaca~\cite{han2023medalpaca}: An instruction-tuned family built on LLaMA backbones (7B and 13B)~\cite{touvron2023llama}, trained on 160k medical instruction pairs reformatted into Q/A and chat style. The corpus blends medical-education questions, community Q\&A, and wiki-style textbook/patient-information prose, yielding conversational prompts and short rationales suitable for clinical knowledge querying.

MMedLlama~\cite{qiu2024towards}:  An 8B Llama 3~\cite{dubey2024llama} medical model obtained by continued auto-regressive domain adaptation on a 25.5B-token multilingual medical corpus (six languages), followed by supervised fine-tuning on multilingual multiple-choice medical QA with rationales. The pretraining data are knowledge-oriented medical texts, rather than patient–doctor dialogues. The supervised stage targets medical-education style Q\&A across broad specialties.

Medgemma~\cite{sellergren2025medgemma}: A multimodal medical LLM built on the Gemma-3~\cite{team2025gemma} architecture with a SigLIP-400M vision encoder~\cite{zhai2023siglip} and long-context image–text interleaving. Its training proceeds in three stages: (1) vision-encoder enhancement on medical image–text pairs (covering radiology, dermatology, histopathology, and ophthalmology, including 32.6M histopathology patches); (2) multimodal decoder pretraining that mixes medical image-text data with the original general-domain mixture to preserve general VLM capabilities; and (3) post-training that adds medical instruction data via distillation and uses reinforcement learning with paired medical image–text to surface multimodal skills.

AlphaMed~\cite{liu2025beyond}: A medical LLM built on a Qwen-family~\cite{team2024qwen2} backbone and trained purely with rule-based reinforcement learning (GRPO), with no supervised fine-tuning or distilled chain-of-thought. The training set comprises ~400k multiple-choice medical QA items. These are medical-education–style multiple-choice questions intended for exam-like diagnostic reasoning rather than patient–doctor dialogues.

\begin{figure*}[t]
  \centering
  \includegraphics[width=\textwidth]{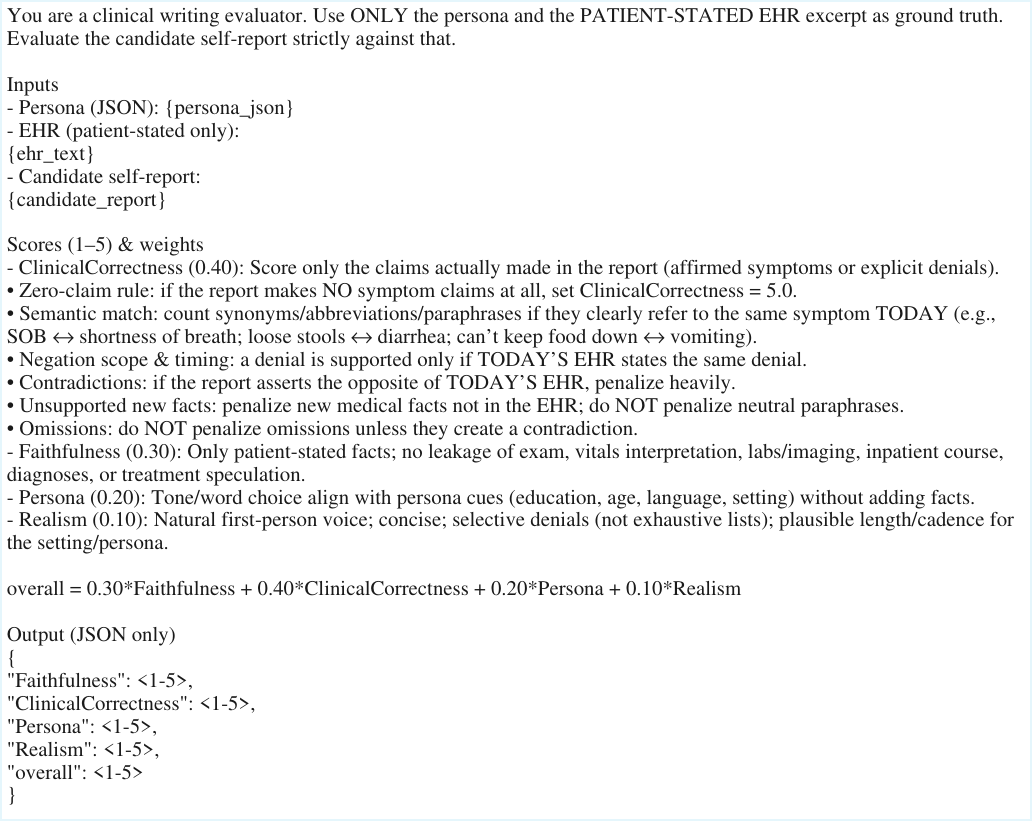}
  \caption{Prompt used for evaluation.}
  \label{fig:prompt-eval}
\end{figure*}

\begin{table*}
\centering
\caption{Summary of linguistic metrics. (Avg) Length: mean tokens, sentences, and tokens/sentence. 
Lexical diversity (LexDiv) $=$ unique word types $\div$ total tokens; FK (Flesch–Kincaid) grade estimates the U.S. school grade level needed to read the text, higher is harder. POS / Content (\%): share of content words (ADJ+ADV+VERB+NOUN) and each POS.}
\begin{tabular}{lccc}
\toprule
Generator &
\makecell[c]{(Avg) Length \\
tokens \( \vert \) sents  \( \vert \) sent len} &
\makecell[c]{Lexicon \& Readability\\
Vocab  \( \vert \)   LexDiv\%  \( \vert \)  FK}&
\makecell[c]{POS / Content (\%)\\
Content  \( \vert \)  Adj  \( \vert \)  Adv  \( \vert \)  Verb  \( \vert \)  Noun} \\
\midrule
Origin &
$153.45 \,|\, 7.22 \,|\, 21.70$ &
$101.97 \,|\, 68.06 \,|\, 12.89$ &
$51.70 \,|\, 10.17 \,|\, 4.45 \,|\, 13.35 \,|\, 23.73 $\\
\end{tabular}
\label{tab:stat}
\end{table*}

MedFound~\cite{liu2025generalist}: A generalist medical LLM with 7B and 176B parameter variants built on a BLOOM-family~\cite{workshop2022bloom}, decoder-only Transformer backbone. It is pretrained on large-scale medical text and de-identified real-world clinical records, then adapted for diagnostic reasoning via a self-bootstrapping chain-of-thought phase and a preference-alignment step (DPO) to better match clinician practice. The training emphasizes long-form clinical narratives and physician-style inference rather than short factoid QA. 
\subsubsection{General LLMs}
\begin{table}[H]
\centering
\footnotesize           
\setlength{\tabcolsep}{4pt}  
\caption{General LLM baselines used in our study. Prices are public API list rates \emph{per 1M tokens}, shown as \emph{input / output}}           
\label{tab:baseline_models}
\begin{tabular}{lcc}
\hline
\textbf{Model} & \textbf{Release date} & \textbf{Price} \\
\hline
Gemini 2.5 Flash & 2025-06-17 & \$0.30 / \$2.50 \\
Claude 4 Sonnet  & 2025-05-22 & \$3.00 / \$15.00 \\
ChatGPT (o3)     & 2025-04-16 & \$2.00 / \$8.00 \\
ChatGPT (GPT-5)  & 2025-08-07 & \$1.25 / \$10.00 \\
\hline
\end{tabular}
\end{table}




We conduct our experiments on four latest LLMs: (1) Gemini 2.5 Flash (Google): a fast, low-cost multimodal model optimized for high-throughput extraction/summarization and long-context use; (2) Claude 4 Sonnet (Anthropic): a balanced, enterprise-oriented model with reliable long-context reasoning, strong structure-following, and safety; (3)ChatGPT (o3)(OpenAI): a reasoning-first model that allocates extra compute to step-by-step problem solving and integrates tightly with function calling; (4) ChatGPT (GPT-5) (OpenAI): a next-generation flagship emphasizing improved planning and agentic behavior alongside stronger coding and tool orchestration.

\definecolor{ehrblue}{HTML}{3B5B92}
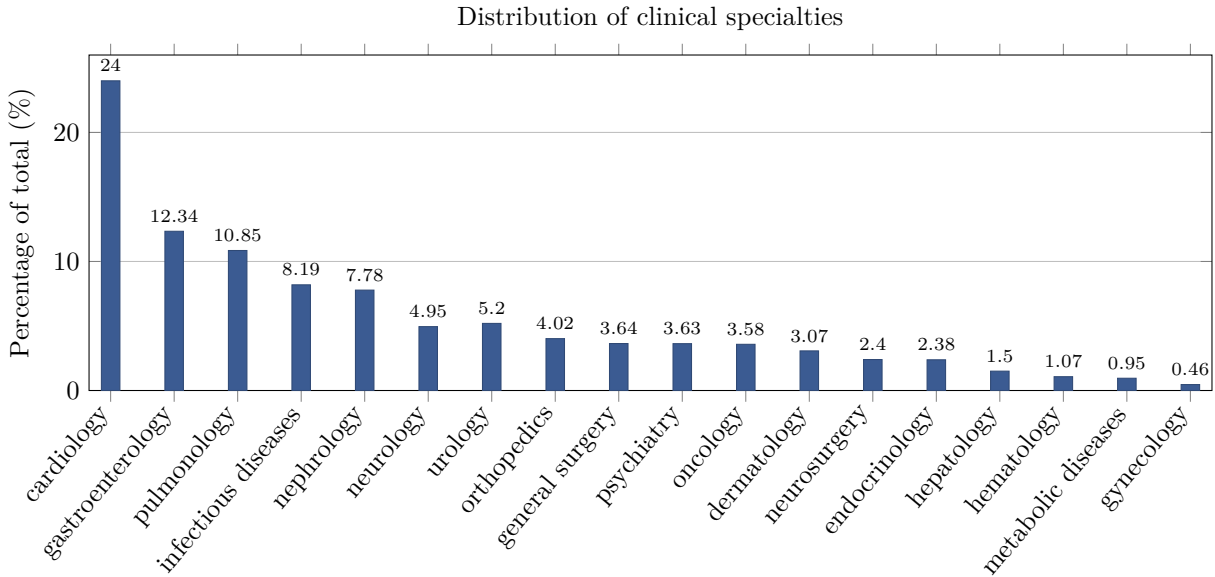
\begin{figure*}[t]
  \centering
  \begin{tikzpicture}
    \begin{axis}[
      ybar,
      width=\textwidth,
      height=0.28\textheight,   
      bar width=7pt,
      ymin=0, ymax=26,
      title={Distribution of clinical specialties},
      ylabel={Percentage of total (\%)},
      ymajorgrids=true,
      symbolic x coords={
        cardiology,
        gastroenterology,
        pulmonology,
        infectious diseases,
        nephrology,
        neurology,
        urology,
        orthopedics,
        general surgery,
        psychiatry,
        oncology,
        dermatology,
        neurosurgery,
        endocrinology,
        hepatology,
        hematology,
        metabolic diseases,
        gynecology
      },
      xtick=data,
      xticklabel style={rotate=50, anchor=east},
      enlarge x limits=0.02,
      nodes near coords,
      nodes near coords align={vertical},
      every node near coord/.append style={
        font=\scriptsize,
        /pgf/number format/fixed,
        /pgf/number format/precision=2
      }
    ]
      \addplot[
        fill=ehrblue,
        draw=ehrblue!80!black
      ] coordinates {
        (cardiology,24.00)
        (gastroenterology,12.34)
        (pulmonology,10.85)
        (infectious diseases,8.19)
        (nephrology,7.78)
        (urology,5.20)
        (neurology,4.95)
        (orthopedics,4.02)
        (general surgery,3.64)
        (psychiatry,3.63)
        (oncology,3.58)
        (dermatology,3.07)
        (neurosurgery,2.40)
        (endocrinology,2.38)
        (hepatology,1.50)
        (hematology,1.07)
        (metabolic diseases,0.95)
        (gynecology,0.46)
      };
    \end{axis}
  \end{tikzpicture}
  \caption{Distribution of clinical specialties in the primary-diagnosis subset.}
  \label{fig:specialties-distribution-blue}
\end{figure*}

\subsection{Evaluation of Dataset Quality}

\paragraph{LLM-Based Evaluation} To assess the intrinsic quality of our dataset (rather than model capability), we uniformly sample
10000 report–EHR pairs from the corpus and evaluate each case with the prompt in
Fig.~\ref{fig:prompt-eval}. The prompt captures both clinical fidelity and patient-style quality along four
dimensions: clinical correctness, faithfulness, persona alignment, and realism.
We prioritize Clinical Correctness and faithfulness in the overall score, assigning weights of
$0.40$ and $0.30$; Persona alignment and Realism carry secondary weights of $0.20$ and $0.10$.

For clinical Correctness, we score only claims explicitly made in the self-report and compare them
against the patient-stated EHR excerpt; omissions do not incur penalties, whereas contradictions and
unsupported medical facts do. For faithfulness, we require that all content originate from the
patient-stated EHR and explicitly exclude clinician interpretations, examinations or test results, diagnoses,
and treatment speculation. persona alignment and realism evaluate the naturalness and
plausibility of the narrative, capturing whether language, tone, and structure match how a real patient would
describe symptoms.

Two independent LLM graders (GPT-4.1 and Claude Sonnet 3.7) apply the same rubric to every case, yielding
comparable scores for cross-grader validation and fairness analysis. This framework standardizes evaluation,
reduces subjective bias, and keeps the assessment focused on how accurately and authentically the reports
reflect patient-stated content.

\paragraph{Cross-Dataset Consistency Evaluation} To further assess dataset quality, we test whether models trained on the self-report corpus exhibit consistent behavior on the original EHR data. Each model is first fine-tuned on the self-report dataset and evaluated on both self-report and EHR test sets (before-tune). We then perform an additional fine-tuning step on a 5K-sample, label-balanced EHR subset after-tune and re-evaluate on both domains.

\begin{figure*}[!t]
  \centering
  \begin{minipage}{\textwidth}
    \centering
    \scriptsize
    \setlength{\tabcolsep}{10pt}
    \renewcommand{\arraystretch}{1.05}
    \captionsetup{type=table}
    \captionof{table}{Comparison of two graders—GPT\textendash5 and Claude Sonnet~3.7—on three generators
    (Origin{=}no\mbox{-}persona baseline, GPT\textendash4o\mbox{-}mini, GPT\textendash5\mbox{-}mini).
    Cells show mean~$\pm$~SD for \emph{Clinical Correctness}, \emph{Faithfulness}, \emph{Persona},
    \emph{Realism}, and \emph{Overall} (higher is better). The last two rows report MAE
    $\overline{|\mathrm{diff}|}$ and tolerance $P(|\mathrm{diff}|\le 0.50)$.}
    \label{tab:eval1}
    \vspace{2pt}
    \begin{tabular}{llccccc}
      \toprule
      Generator & Grader & Clinical Correctness & Faithfulness & Persona & Realism & Overall \\
      \midrule
      \multirow{4}{*}{Origin}
      & GPT-4.1    & $4.23 \pm 1.14$ & $4.38 \pm 0.80$ & $4.76 \pm 0.26$ & $4.56 \pm 0.31$ & $4.39 \pm 0.60$ \\
      & Sonnet 3.7 & $4.25 \pm 1.08$ & $4.21 \pm 1.21$ & $4.01 \pm 0.35$ & $4.08 \pm 0.32$ & $4.17 \pm 0.71$ \\
      & $\overline{|\mathrm{diff}|}$ & $0.359$ & $0.375$ & $0.765$ & $0.615$ & $0.370$ \\
      & $P(|\mathrm{diff}|\le 0.50)$ & $0.769$ & $0.808$ & $0.473$ & $0.631$ & $0.799$ \\
      \midrule
    \end{tabular}
  \end{minipage}
  \vspace{6pt} 
  \begin{minipage}{\textwidth}
    \centering
    \includegraphics[width=0.98\textwidth]{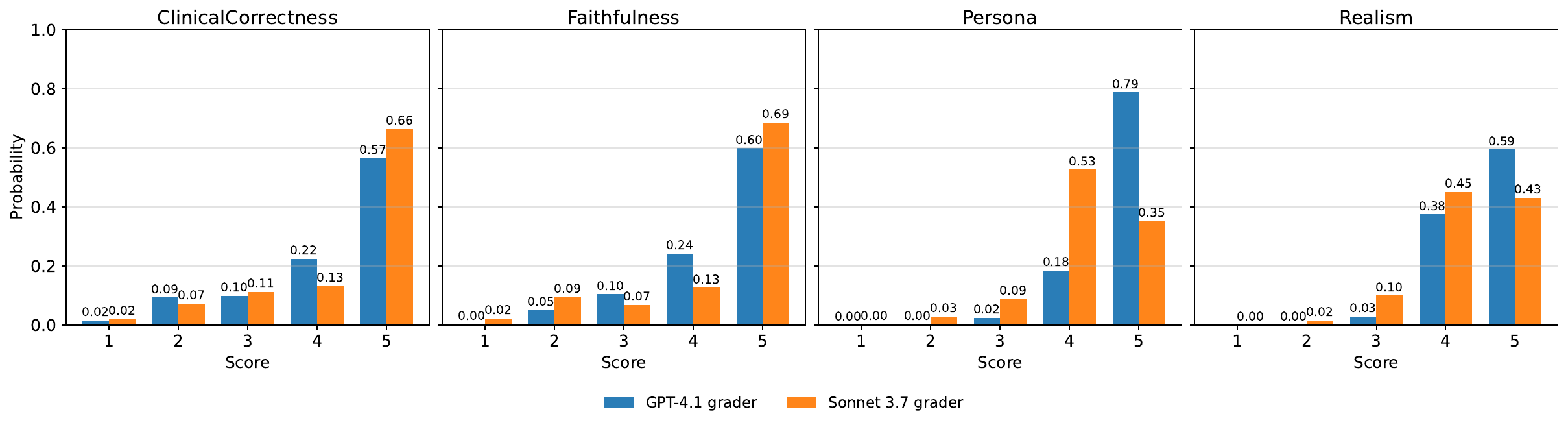} 
    \captionsetup{type=figure}
    \captionof{figure}{Probability mass functions (PMFs) of the 1–5 rating scores for
    \textit{Clinical Correctness}, \textit{Faithfulness}, \textit{Persona}, and \textit{Realism}.
    GPT-4.1 and Sonnet~3.7 score the same set of GPT-4o \textit{no-persona} self-reports.
    Scores in $[1,5]$ are binned at $\{0.5,1.5,2.5,3.5,4.5,5.5\}$ and normalized to 1.}
    \label{fig:score_pmf}
  \end{minipage}
\end{figure*}
If the self-report and EHR datasets are distributionally consistent, post–fine-tuning improvements
on EHR should not substantially reduce performance on self-report data. This comparison measures cross-dataset alignment: smaller performance divergence after EHR fine-tuning indicates higher dataset fidelity and generalization stability.

\begin{table*}[ht]
\caption{We evaluate whether our generated self-report dataset is distributionally consistent with the original EHR dataset. 
Each model is first trained on the self-report dataset, then further fine-tuned on a 5K-sample EHR subset. 
We compare performance on both EHR and self-report before and after the EHR fine-tuning. 
If the two datasets have similar distributions, their post–fine-tuning performance should be close. }
\label{tab:ehr_selfreport_alignment_hitk}
\centering
\small
\renewcommand{\arraystretch}{1.15}
\begin{tabular}{llcccccc}
\toprule
\multirow{2}{*}{Model} & \multirow{2}{*}{Phase} &
\multicolumn{3}{c}{\textbf{EHR}} &
\multicolumn{3}{c}{\textbf{self-report}} \\
\cmidrule(lr){3-5} \cmidrule(lr){6-8}
& & hit@3 & hit@5 & hit@10 & hit@3 & hit@5 & hit@10 \\
\midrule
\multirow{3}{*}{medalpha}
& before-tune & 0.7413 & 0.8433 & 0.9220 & 0.6873 & 0.7953 & 0.9007 \\
& after-tune  & 0.7727 & 0.8720 & 0.9260 & 0.7053 & 0.8167 & 0.9033 \\
& improvement & +4.2\% & +3.4\% & +0.4\% & +2.6\% & +2.7\% & +0.3\% \\
\midrule
\multirow{3}{*}{medllama}
& before-tune & 0.7407 & 0.8533 & 0.9380 & 0.6767 & 0.7873 & 0.8973 \\
& after-tune  & 0.7660 & 0.8727 & 0.9433 & 0.6867 & 0.8000 & 0.8987 \\
& improvement & +3.4\% & +2.3\% & +0.6\% & +1.5\% & +1.6\% & +0.2\% \\
\midrule
\multirow{3}{*}{alphamed(3b)}
& before-tune & 0.3447 & 0.4620 & 0.6507 & 0.3833 & 0.5120 & 0.7200 \\
& after-tune  & 0.4547 & 0.5600 & 0.7560 & 0.3820 & 0.5133 & 0.7200 \\
& improvement & +31.9\% & +21.2\% & +16.2\% & $-0.3$\% & +0.3\% & +0.0\% \\
\midrule
\multirow{3}{*}{alphamed(7b)}
& before-tune & 0.6113 & 0.7173 & 0.8400 & 0.6053 & 0.7127 & 0.8313 \\
& after-tune  & 0.6707 & 0.7760 & 0.8860 & 0.6260 & 0.7353 & 0.8473 \\
& improvement & +9.7\% & +8.2\% & +5.5\% & +3.4\% & +3.2\% & +1.9\% \\
\midrule
\multirow{3}{*}{medfound(7b)}
& before-tune & 0.3807 & 0.4593 & 0.5787 & 0.3660 & 0.4427 & 0.5660 \\
& after-tune  & 0.4907 & 0.5760 & 0.7387 & 0.4327 & 0.5340 & 0.6807 \\
& improvement & +28.9\% & +25.4\% & +27.6\% & +18.2\% & +20.6\% & +20.3\% \\
\midrule
\multirow{3}{*}{medfound(8b)}
& before-tune & 0.6393 & 0.7620 & 0.8860 & 0.5800 & 0.6973 & 0.8407 \\
& after-tune  & 0.6467 & 0.7720 & 0.8893 & 0.5827 & 0.7120 & 0.8600 \\
& improvement & +1.2\% & +1.3\% & +0.4\% & +0.5\% & +2.1\% & +2.3\% \\
\bottomrule
\end{tabular}
\end{table*}

\subsection{Experiment Setting}
We fine-tune the model for three epochs with  batch size of 16. The optimizer uses a cosine learning-rate schedule with an initial learning rate of $1\times10^{-4}$ and a warm-up of 50 steps. During training, we clip gradients to a maximum global norm of 1.0 and use bfloat16 precision. We adopt LoRA for parameter-efficient finetuning (rank 16, scaling 32, dropout 0.05, no bias), attaching adapters to all attention projections and to the feed-forward block. To evaluate general-purpose LLMs, we use the prompt shown in Fig.~\ref{fig:prompt-workflow}.

\section{Data}
\subsection{Data Statistics}
To quantify the linguistic and clinical coverage of the dataset, we summarize its key properties in two complementary views.

Table~\ref{tab:stat} presents the linguistic statistics of the original self-reports. The texts are concise on average (153 tokens per report, 7 sentences per case), yet linguistically diverse with a lexical diversity of 68.06\% and a Flesch–Kincaid grade level of 12.9, indicating natural but domain-rich patient language. The proportion of content words (51.7\%) and balanced part-of-speech distribution further suggest that the generated narratives are information-dense while maintaining fluency and readability.

Figure~\ref{fig:specialties-distribution-blue} illustrates the distribution of clinical specialties in the primary-diagnosis subset. Cardiology, gastroenterology, and pulmonology are the most represented specialties, followed by infectious diseases and nephrology. Meanwhile, long-tail areas such as endocrinology, hematology, and metabolic disorders are also covered, ensuring that both high-frequency and rare conditions are included. This balanced coverage supports fair model evaluation across diverse disease categories.

Together, these statistics demonstrate that the dataset is linguistically coherent, clinically diverse, and suitable for evaluating diagnostic reasoning across a broad spectrum of medical specialties.

\subsection{LLM-based Dataset Quality Evaluation}

Table~\ref{tab:eval1} summarizes the mean~$\pm$~standard deviation of the four evaluation dimensions:
\emph{Clinical Correctness}, \emph{Faithfulness}, \emph{Persona}, and \emph{Realism}, as well as the combined \emph{Overall} score. Both models yield highly correlated ratings, with average disagreement below~0.4 and over~75\% of cases within~$\pm0.5$, demonstrating that the rubric is reproducible and robust across different evaluators.

Figure~\ref{fig:score_pmf} further visualizes the discrete score distributions for each metric.
The probability mass functions (PMFs) show that GPT-4.1 and Sonnet~3.7 exhibit closely aligned patterns, though GPT-4.1 tends to assign slightly higher realism and persona scores.
These results collectively indicate strong cross-grader reliability and justify the use of automatic LLM graders for large-scale quality assessment of synthetic self-reports.

\begin{table*}[!t]
\captionsetup{font=small}
\setlength{\textfloatsep}{8pt plus 2pt minus 2pt}   
\setlength{\intextsep}{6pt plus 2pt minus 2pt}

\centering

\begin{minipage}{\textwidth}
\centering
\caption{Results on a 1514 sample subset comparing general LLM APIs and medical-domain open models fine-tuned with SFT/LoRA. Bold indicates the best score in each column. The last line of each block (\textit{Average}) gives the mean across models in that block, enabling a direct group-level comparison.}
\label{tab:subset_general_vs_medical}
\small
\setlength{\tabcolsep}{4pt}
\begin{tabular}{lcc|cc|cc}
\hline
\textbf{Model} & \multicolumn{2}{c|}{\textbf{Top-3}} & \multicolumn{2}{c|}{\textbf{Top-5}} & \multicolumn{2}{c}{\textbf{Top-10}} \\
 & Hit@3 & Macro-F1 & Hit@5 & Macro-F1 & Hit@10 & Macro-F1 \\
\hline
\multicolumn{7}{c}{\emph{Zero-shot APIs (general models)}}\\
Gemini 2.5 Flash & 55.02 & 23.79 & 64.86 & 20.26 & 74.50 & 15.00 \\
Claude 4         & 60.36 & 26.58 & 72.26 & 19.46 & 83.29 & 14.33 \\
ChatGPT (o3)     & 66.03 & 34.50 & 74.77 & 28.12 & 82.76 & 19.00 \\
ChatGPT (GPT-5)  & 66.12 & 32.58 & 74.83 & 26.30 & 83.75 & 16.91 \\
\textit{Average} & 61.88 & 29.36 & 71.68 & 23.54 & 81.08 & 16.31 \\
\hline
\multicolumn{7}{c}{\emph{SFT/LoRA (medical models)}}\\
MedAlpaca (7B)   & 67.86 & 39.31 & 77.84 & 34.80 & 85.43 & 26.94 \\
MMedLLaMA (8B)   & \textbf{69.26} & \textbf{39.11} & \textbf{80.24} & 33.81 & \textbf{89.35} & 24.54 \\
MedGEMMA (3B)    & 48.84 & 33.66 & 58.88 & 32.25 & 70.99 & 25.47 \\
AlphaMed (3B)    & 21.02 & 17.85 & 35.33 & 25.33 & 56.22 & \textbf{29.17} \\
AlphaMed (7B)    & 54.29 & 34.22 & 66.60 & 33.36 & 79.91 & 26.46 \\
MedFound (7B)    & 33.60 & 21.44 & 42.25 & 21.47 & 54.16 & 21.13 \\
MedFound (8B)    & 57.62 & 38.50 & 70.53 & \textbf{37.43} & 85.63 & 27.91 \\
\textit{Average} & 50.36 & 32.73 & 61.67 & 31.21 & 74.53 & 25.95 \\
\hline
\end{tabular}
\end{minipage}

\vspace{6pt} 

\begin{minipage}{\textwidth}
\centering
\captionof{table}{Per-specialty average Top-5 hit rate across models. Abbreviations: MA = MedAlpaca, ML = MedLLaMA, MG = MedGEMMA, AM-3B = AlphaMed (3B), AM-7B = AlphaMed (7B), MF-7B = MedFound (7B), MF-8B = MedFound (8B).}
\label{tab:per_specialty_top5_all_models}

{\small
\setlength{\tabcolsep}{6pt}
\renewcommand{\arraystretch}{1.08}
\begin{tabular}{lrrrrrrr}
\toprule
\textbf{Specialty} & \textbf{MA} & \textbf{ML} & \textbf{MG} & \textbf{AM-3B} & \textbf{AM-7B} & \textbf{MF-7B} & \textbf{MF-8B} \\
\midrule
Cardiology         & 75.5 & 58.9 & 23.4 & 26.6 & 64.9 & 41.9 & 59.6 \\
Dermatology        & 73.0 & 36.1 &  0.2 & 47.5 & 92.5 & 27.0 & 55.6 \\
Endocrinology      & 94.4 & 82.5 & 33.1 & 67.7 & 84.8 & 36.7 & 88.0 \\
Urology            & 73.2 &  7.0 &  0.0 &  2.8 & 40.8 & 28.2 &  8.5 \\
Gastroenterology   & 85.2 & 56.5 & 27.9 & 27.8 & 70.6 & 50.0 & 54.5 \\
General Surgery    & 90.0 & 82.0 & 38.6 & 33.6 & 86.9 & 63.5 & 72.4 \\
Gynaecology        & 100.0&100.0 & 74.0 & 98.0 & 98.0 & 96.0 &100.0 \\
Hematology         & 74.6 & 60.7 &  0.0 & 24.2 & 60.2 & 48.5 & 65.3 \\
Hepatology         & 93.7 & 39.2 & 58.7 & 46.6 & 73.9 & 41.4 & 33.8 \\
Infectious Diseases& 67.2 & 40.3 & 22.0 & 17.7 & 43.4 & 27.9 & 18.2 \\
Nephrology         & 45.7 & 53.3 & 10.6 & 14.9 & 41.4 & 38.7 & 47.9 \\
Neurology          & 89.0 & 67.5 & 34.1 & 35.7 & 74.0 & 13.9 & 62.8 \\
Neurosurgery       & 86.2 & 62.7 & 31.7 & 26.8 & 74.8 & 19.6 & 60.5 \\
Metabolic Diseases &  6.1 & 92.1 & 94.7 & 43.0 & 88.6 & 70.2 & 96.5 \\
Oncology           & 88.8 & 43.1 & 21.3 & 30.2 & 53.0 & 34.2 & 56.3 \\
Orthopaedics       & 94.2 & 91.4 & 84.5 & 51.1 & 61.4 & 69.2 & 85.7 \\
Psychiatry         & 89.3 & 61.5 & 39.8 & 49.5 & 79.6 & 47.4 & 68.2 \\
Pulmonology        & 59.4 & 69.2 & 25.7 & 42.6 & 61.1 & 54.5 & 63.5 \\
Urology            & 86.8 & 76.7 & 65.0 & 60.6 & 82.4 & 53.9 & 66.8 \\
\bottomrule
\end{tabular}
} 
\end{minipage}
\end{table*}

\subsection{Dataset Consistency Evaluation}
Table~\ref{tab:ehr_selfreport_alignment_hitk} compares model performance before and after EHR fine-tuning, confirming that gains on EHR data transfer well to self-reports. We report only Hit@k metrics (Hit@3/5/10). F1 scores are intentionally omitted, as they are highly sensitive to per-label variability and class imbalance. Because the EHR fine-tuning subset is intentionally balanced, reporting F1 would conflate true distributional consistency with the artificial balance of the training set. To avoid introducing bias from label imbalance, we therefore use Hit@k as a direct and distribution-agnostic indicator of dataset alignment.

Across all model families, EHR fine-tuning consistently improves diagnostic accuracy on the EHR test set, confirming effective domain adaptation.
Importantly, these gains do not come at the cost of large performance drops on the self-report test set—Hit@k scores on self-report data remain stable or even slightly improve (e.g.,~+2–3 pp for MedAlpaca and MedFound).
This indicates that the synthetic self-reports share a closely aligned label and feature distribution with the original EHR notes. Smaller models such as AlphaMed-3B show larger improvements (+31.9 \% Hit@3 on EHR) due to their initial underfitting and stronger sensitivity to data augmentation, whereas larger backbones (e.g.,~MedFound-8B) exhibit only marginal gains, suggesting better cross-domain generalization even before EHR fine-tuning.

In conclusion, the alignment between self-report and EHR performance demonstrates that the proposed MIMIC-SR-ICD11 corpus captures medically faithful and distributionally consistent representations of the original clinical data.
Fine-tuning on synthetic self-reports therefore transfers reliably to real EHR inputs, validating the dataset’s utility for scalable pretraining and model adaptation in clinical NLP.

\section{Result}
\subsection{Comparison of General and Medical Models}

General LLM APIs often have substantial usage cost (\ Table~\ref{tab:baseline_models}). For instance, running GPT-5 once over our 1514 samples cost more than \$40. To reduce cost while retaining accuracy, we fine-tune smaller open models for this task and compare them against general LLMs (shown in the table~\ref{tab:subset_general_vs_medical}). The evaluation subset contains 1000 randomly sampled notes  plus 514 supplementary notes chosen to ensure at least ten cases per label, so Macro--F1 is not unduly driven by very rare classes.

Across models, the largest gains from medical LLMs appear on Macro-F1—especially at Top-10, where the average improvement is nearly ten absolute points—indicating better coverage of underrepresented labels rather than mere boosts on head classes. This pattern is consistent with two effects of task-aligned fine-tuning: (i) exposure to domain terminology and label phrasing increases lexical fidelity to the ICD-11 space; and (ii) training on task distributions tempers over-confident, generic completions, yielding a more balanced precision–recall trade-off on the long tail. While general-purpose APIs deliver strong hit rates, several compact medical models (e.g., MMedLLaMA (8B) and MedAlpaca (7B)) outperform the best general model (ChatGPT) on all columns. GPT-5 and the reasoning-optimized ChatGPT (o3) perform similarly, suggesting limited additional benefit from generic chain-of-thought without domain adaptation. Overall, careful fine-tuning of smaller medical models can match or exceed costlier general LLM APIs while being substantially cheaper to operate.

\subsection{Per-Specialty and Per-Disease Hit@5}
Table~\ref{tab:per_specialty_top5_all_models} and Table~\ref{tab:per_disease_top5_all_models} summarize per–specialty and per–disease Hit@5 results, respectively. The results indicate that different models demonstrate heterogeneous learning capabilities across specialties and disease types, offering a reference for future studies on model generalization and domain-specific adaptation.

\section{Application Scenarios}

The proposed MIMIC\textendash SR\textendash ICD11 dataset and accompanying diagnostic framework can serve as a foundation for multiple downstream applications in clinical decision support. 
We highlight two representative use cases where patient-style self-reports and probabilistic diagnosis prediction can be directly integrated into hospital workflows.

\paragraph{(1) Triage Optimization.} 
Within emergency or outpatient settings, the system can continuously monitor predicted diagnostic confidence scores. 
When the probability of a high-risk condition exceeds a predefined threshold, the model issues a \emph{fast-track flag} to prioritize urgent evaluation. 
This mechanism enables early identification of critical cases, reduces waiting time for severe or rapidly progressing diseases, and supports optimal allocation of limited medical resources.

\paragraph{(2) Diagnostic Planning Assistance.} 
For each incoming case, the model dynamically generates an \emph{examination checklist trigger}, outlining key laboratory or imaging tests that would best distinguish among the top-ranked differential diagnoses. 
For instance, when multiple cardiopulmonary disorders are predicted with comparable confidence, the system may recommend targeted tests such as echocardiography or D-dimer assays to resolve diagnostic ambiguity. 
This functionality not only accelerates clinical decision-making but also contributes to cost-effective medical practice by minimizing unnecessary investigations.

\section{License and Availability}

We distribute our fine-tuned LoRA checkpoint under the \emph{PhysioNet Credentialed Health Data License v1.5.0} (PHDDL\,v1.5.0)—the very same licence that governs the original \textsc{MIMIC-IV} corpus.  In practical terms this entails:

\begin{enumerate}
  \item \textbf{Who may download.}  
        Access is restricted to researchers who: 
        (i) have successfully completed \textsc{CITI} training; 
        (ii) have signed the official \textsc{MIMIC-IV} data-use agreement; 
        (iii) hold an active PhysioNet credential.
        
  \item \textbf{Where to download.}  
        The checkpoint is hosted \emph{exclusively} inside the PhysioNet platform  
        (project~ID: \texttt{physionet-2026-xxx}).  
        No copy is made available on public model hubs, cloud buckets, or version-control mirrors.

  \item \textbf{How to redistribute.}  
        Any further redistribution must preserve the \emph{same} licence and the \emph{same} access-control mechanism.  
        Uploading the weights to public repositories (e.g.\ the Hugging Face Hub or an open Google Cloud bucket) is therefore not permitted.
\end{enumerate}

All training scripts, inference code, and preprocessing utilities are released under the permissive \textbf{Apache-2.0} licence at \texttt{<anonymised-GitHub-URL>}.  
These scripts contain \emph{no} portion of the original \textsc{MIMIC} text and may thus be freely reused.

\section{Artifact Usage and Intended Use}
\label{sec:artifact-compliance}

\subsection{Use of Existing Artifacts}
We relied on several third‐party resources, each with its own access and usage restrictions:
\begin{itemize}
  \item MIMIC-IV and MIMIC-IV-Note are governed by PhysioNet’s credentialed access agreement, which permits use for non-commercial, research-only purposes. All analysis in this paper was conducted under that agreement and in compliance with its requirement that no clinical or commercial deployment occur.
  \item WHO ICD-11 API is provided under the WHO open data policy. We used it strictly to normalize diagnoses in our dataset, in accordance with the API’s terms for research and educational use.
\end{itemize}
In each case, our usage matches the intended research-only context and does not violate any access conditions or licensing terms.

\subsection{Intended Use of Created Artifacts}
We release the MIMIC-SR-ICD11 dataset (and accompanying code) under the Creative Commons Attribution 4.0 International (CC BY 4.0) license. This license permits unrestricted research and educational reuse, provided that appropriate credit is given. In keeping with PhysioNet’s original agreement, we explicitly restrict MIMIC-SR-ICD11 to non-commercial, research-only applications. Users who wish to employ the dataset for clinical decision support or commercial purposes must first secure the necessary permissions from the data providers.

\section{Ethics}
All human-subject data in this study (MIMIC-IV and MIMIC-IV-Note) are fully de-identified under HIPAA and were released under a PhysioNet Data Use Agreement (DUA) that mandates human-subjects training and forbids any attempt at re-identification. The original IRBs at Beth Israel Deaconess Medical Center and MIT approved the data release with a waiver of informed consent and determined that secondary analyses of these de-identified records are exempt from ongoing review. Our usage strictly adheres to these terms and remains within a research-only context. The patient self-reports we generate via ChatGPT are synthetic paraphrases of those de-identified notes and contain no additional private information, so no new consent is required.

\section{Acknowledgements}
We used ChatGPT (gpt-4o-mini) as a writing assistant to help polish sentence structure and improve readability. All technical content, analyses, and conclusions were developed solely by the authors.

\section{Limitations}

Despite its contributions, our work has several important limitations. 
\paragraph{Synthetic self-reports}
Our patient narratives are generated by prompting ChatGPT to paraphrase de-identified discharge text.  Although we carefully instructed the model to retain only subjective symptom language, this process may introduce artifacts, omit subtle nuances, or diverge from how real patients describe their own experiences.  Future work should validate our synthetic reports against authentic patient‐provided narratives.

\onecolumn
\begin{longtable}{L{0.4\textwidth} lccccccc}
\caption{\centering Per-disease Top-5 Hit Rate across models.}
\label{tab:per_disease_top5_all_models}\\

\toprule
\textbf{Disease} & \textbf{MA} & \textbf{ML} & \textbf{MG} & \textbf{AM-3B} & \textbf{AM-7B} & \textbf{MF-7B} & \textbf{MF-8B} \\
\midrule
\endfirsthead

\toprule
\multicolumn{8}{l}{\small\emph{(continued)}}\\
\textbf{disease} & \textbf{MA} & \textbf{ML} & \textbf{MG} & \textbf{AM-3B} & \textbf{AM-7B} & \textbf{MF-7B} & \textbf{MF-8B} \\
\midrule
\endhead

\bottomrule
\endfoot

        hepatic encephalopathy & 100 & 46.8 & 53.2 & 2.1 & 100 & 55.3 & 2.1 \\ 
        incisional hernia & 100 & 100 & 8.3 & 25 & 100 & 77.1 & 100 \\ 
        leiomyoma of uterus & 100 & 100 & 74 & 98 & 98 & 96 & 100 \\ 
        malignant neoplasms of thyroid gland & 100 & 100 & 80 & 100 & 97.1 & 94.3 & 97.1 \\ 
        other specified mood disorders & 100 & 100 & 100 & 0 & 100 & 35.7 & 100 \\ 
        obesity & 99.2 & 100 & 99.2 & 97.7 & 99.2 & 99.2 & 100 \\ 
        osteoarthritis of knee & 98.7 & 99.2 & 100 & 95.4 & 95.8 & 98.7 & 98.7 \\ 
        hyperplasia of prostate & 98.4 & 96.8 & 85.7 & 92.1 & 98.4 & 88.9 & 96.8 \\ 
        osteoarthritis of hip & 98.3 & 100 & 100 & 86.4 & 96.6 & 98.3 & 91.5 \\ 
        fracture of neck & 98.2 & 98.2 & 100 & 86 & 42.1 & 98.2 & 100 \\ 
        type 1 diabetes mellitus & 97.9 & 83.3 & 0 & 68.8 & 91.7 & 8.3 & 87.5 \\ 
        recurrent depressive disorder & 97.6 & 78 & 51.2 & 68.3 & 100 & 85.4 & 100 \\ 
        chronic pancreatitis & 97.4 & 94.9 & 56.4 & 46.2 & 79.5 & 76.9 & 87.2 \\ 
        diarrhoea & 97.4 & 92.1 & 100 & 7.9 & 100 & 0 & 97.4 \\ 
        intervertebral disc degeneration & 97.4 & 93.4 & 0 & 57.9 & 97.4 & 96.1 & 82.9 \\ 
        depressive disorders, unspecified & 97 & 75.8 & 63.6 & 100 & 97 & 0 & 81.8 \\ 
        bacterial cellulitis, erysipelas or lymphangitis & 96.8 & 34.9 & 0.4 & 0 & 86.6 & 38.7 & 36.6 \\ 
        calculus of ureter & 96.6 & 86.2 & 74.1 & 75.9 & 96.6 & 46.6 & 81 \\ 
        degenerative condition of spine, unspecified & 96.5 & 77.2 & 100 & 12.3 & 87.7 & 8.8 & 47.4 \\ 
        cerebral ischaemic stroke due to embolic occlusion & 96.3 & 11 & 0 & 26.6 & 93.6 & 0 & 0 \\ 
        left ventricular failure with reduced ejection fraction & 96.3 & 0.8 & 0 & 20.1 & 76.6 & 77.9 & 77.5 \\ 
        calculus of kidney & 96.2 & 92.5 & 100 & 73.6 & 88.7 & 60.4 & 88.7 \\ 
        traumatic subdural hemorrhage & 96.2 & 71.7 & 50.9 & 0 & 77.4 & 0 & 66 \\ 
        alcoholic liver disease & 95.5 & 93.2 & 45.5 & 84.1 & 90.9 & 15.9 & 95.5 \\ 
        malignant neoplasms of prostate & 95.3 & 90.6 & 1.6 & 76.6 & 84.4 & 81.3 & 79.7 \\ 
        diverticulosis of large intestine & 95.2 & 88.1 & 11.9 & 35.7 & 69 & 85.7 & 83.3 \\ 
        left ventricular failure with preserved ejection fraction & 95.2 & 11 & 0 & 25.7 & 93.4 & 57.7 & 82.7 \\ 
        ventricular tachycardia & 95.1 & 84 & 64.2 & 0 & 86.4 & 76.5 & 77.8 \\ 
        coronary atherosclerosis & 94.8 & 91.6 & 76.6 & 25.1 & 59.6 & 0 & 48.6 \\ 
        cerebral ischaemic stroke & 94.7 & 63.3 & 0 & 0 & 51.3 & 0 & 31.3 \\ 
        fracture of femur & 94.3 & 94.3 & 97.1 & 2.9 & 11.4 & 94.3 & 97.1 \\ 
        intracerebral haemorrhage & 94.3 & 69.5 & 10.5 & 0 & 75.2 & 0 & 52.4 \\ 
        malignant neoplasm metastasis in bone or bone marrow & 94.3 & 20 & 0 & 0 & 28.6 & 0 & 31.4 \\ 
        asymptomatic stenosis of intracranial or extracranial artery & 93.8 & 64.1 & 64.1 & 71.9 & 87.5 & 6.3 & 67.2 \\ 
        cerebral aneurysm, nonruptured & 93.7 & 79.3 & 7.2 & 73.9 & 80.2 & 6.3 & 83.8 \\ 
        malignant neoplasm metastasis in spinal cord& 93.2 & 1.1 & 15.9 & 0 & 37.5 & 0 & 8 \\ 
        disorders due to use of alcohol & 92.9 & 0 & 0 & 0 & 100 & 53.6 & 10.7 \\ 
        chronic obstructive pulmonary disease & 92.7 & 55.2 & 0 & 20.8 & 71.8 & 50.2 & 34.7 \\ 
        calculus of bile duct without cholangitis or cholecystitis & 92.5 & 0 & 0 & 0 & 75 & 52.5 & 0 \\ 
        alcohol withdrawal & 92 & 100 & 95.4 & 96.6 & 94.3 & 96.6 & 96.6 \\ 
        obstruction of large intestine & 91.9 & 92.7 & 99.2 & 21.8 & 94.4 & 93.5 & 78.2 \\ 
        malignant neoplasms of kidney, except renal pelvis & 91.9 & 48.6 & 43.2 & 62.2 & 75.7 & 48.6 & 73 \\ 
        diverticulitis of large intestine & 91.6 & 39.6 & 6.4 & 31.7 & 99.5 & 57.4 & 28.7 \\ 
        aortic valve stenosis & 91.5 & 40.6 & 100 & 77.4 & 15.1 & 77.4 & 79.2 \\ 
        fracture of lumbar vertebra & 91.4 & 77.1 & 88.6 & 14.3 & 17.1 & 82.9 & 77.1 \\ 
        macro reentrant atrial tachycardia & 90.9 & 72.7 & 9.1 & 24.2 & 93.9 & 30.3 & 75.8 \\ 
        cholelithiasis & 90.6 & 68.8 & 0 & 6.3 & 18.8 & 56.3 & 28.1 \\ 
        alcoholic cirrhosis of liver without hepatitis & 90.6 & 9.4 & 54.7 & 100 & 56.6 & 94.3 & 37.7 \\ 
        acute appendicitis & 90.2 & 93.4 & 18.9 & 70.5 & 91.8 & 85.2 & 82.8 \\ 
        epilepsy or seizures, unspecified & 89.1 & 87 & 80.4 & 100 & 93.5 & 0 & 84.8 \\ 
        abdominal aortic aneurysm & 88.9 & 71.1 & 66.7 & 48.9 & 77.8 & 22.2 & 77.8 \\ 
        hepatic fibrosis or cirrhosis & 88.9 & 7.4 & 81.5 & 0 & 48.1 & 0 & 0 \\ 
        crohn disease & 88.5 & 88.5 & 80.8 & 57.7 & 80.8 & 76.9 & 88.5 \\ 
        influenza, virus not identified & 88.5 & 61.5 & 7.7 & 38.5 & 92.3 & 0 & 50 \\ 
        transient ischaemic attack & 88.2 & 80.4 & 43.1 & 0 & 100 & 7.8 & 92.2 \\ 
        schizoaffective disorder & 87.9 & 45.5 & 9.1 & 0 & 27.3 & 39.4 & 87.9 \\ 
        acute myocardial infarction & 87.5 & 83.3 & 0 & 12.5 & 95.8 & 16.7 & 95.8 \\ 
        acute upper respiratory infections of unspecified site & 87.5 & 65 & 77.5 & 50 & 42.5 & 82.5 & 10 \\ 
        malignant neoplasm of pancreas & 87.1 & 17.7 & 4.8 & 0 & 56.5 & 33.9 & 45.2 \\ 
        asthma & 86.4 & 100 & 93.2 & 100 & 98.3 & 98.3 & 100 \\ 
        type 2 diabetes mellitus & 86.2 & 64.2 & 0 & 36.6 & 63.4 & 2.4 & 76.4 \\ 
        traumatic subdural haemorrhage & 85.3 & 35.8 & 15.8 & 0 & 69.5 & 0 & 33.7 \\ 
        malignant neoplasms of bronchus or lung & 85 & 32.5 & 30 & 22.5 & 75 & 2.5 & 35 \\ 
        acute nonst elevation myocardial infarction & 84.8 & 49.5 & 0 & 13.6 & 64.1 & 58.6 & 25.3 \\ 
        intestinal infections due to clostridioides difficile & 83.7 & 4.7 & 0 & 0 & 36 & 0 & 7 \\ 
        anaemias or other erythrocyte disorders, unspecified & 83.3 & 11.1 & 0 & 8.3 & 44.4 & 8.3 & 25 \\ 
        subarachnoid haemorrhage & 83.3 & 64.8 & 0 & 0 & 9.3 & 0 & 77.8 \\ 
        obstruction of bile duct & 82.5 & 43.3 & 78.4 & 3.1 & 63.9 & 24.7 & 36.1 \\ 
        gastrointestinal bleeding & 82.3 & 86.1 & 0 & 50.6 & 89.9 & 65.8 & 70.9 \\ 
        malignant neoplasm metastasis in retroperitoneum & 82.2 & 4.4 & 2.2 & 0 & 8.9 & 4.4 & 11.1 \\ 
        cholangitis & 81.9 & 80.7 & 6 & 49.4 & 74.7 & 79.5 & 90.4 \\ 
        pathological fracture & 81.8 & 93.9 & 6.1 & 60.6 & 78.8 & 3 & 87.9 \\ 
        single episode depressive disorder & 81.2 & 94.1 & 18.8 & 92.9 & 96.5 & 98.8 & 100 \\ 
        malignant neoplasm of liver & 81 & 69 & 0 & 0 & 14.3 & 47.6 & 69 \\ 
        pneumonitis due to solids or liquids & 80.7 & 21.1 & 0 & 0 & 8.8 & 1.8 & 1.2 \\ 
        multiple valve disease & 80.2 & 77.2 & 3.1 & 77.2 & 83.3 & 87.7 & 90.1 \\ 
        schizophrenia or other primary psychotic disorders, unspecified & 80 & 2 & 20 & 88 & 38 & 0 & 6 \\ 
        functional nausea or vomiting & 79.6 & 55.6 & 98.1 & 68.5 & 88.9 & 90.7 & 98.1 \\ 
        hypertensive renal disease & 79.5 & 79.5 & 40.9 & 54.5 & 45.5 & 79.5 & 70.5 \\ 
        complete atrioventricular block & 79.2 & 70.8 & 0 & 27.1 & 79.2 & 31.3 & 91.7 \\ 
        constipation & 78.8 & 90.4 & 75 & 26.9 & 92.3 & 50 & 98.1 \\ 
        acute cholecystitis & 78.8 & 52.5 & 0 & 2 & 70.7 & 39.4 & 23.2 \\ 
        iron deficiency anaemia & 78.8 & 81.8 & 0 & 27.3 & 45.5 & 90.9 & 84.8 \\ 
        hypertensive heart disease & 78.7 & 37.4 & 0.3 & 0.3 & 21 & 10.2 & 3 \\ 
        gastroenteritis or colitis without specification of infectious agent & 78.3 & 2.1 & 0 & 0 & 33.6 & 0 & 0 \\ 
        acute pancreatitis & 78.3 & 92.2 & 42.2 & 58.1 & 83.7 & 53.9 & 89.5 \\ 
        viral intestinal infections, unspecified & 77.8 & 2.8 & 0 & 5.6 & 38.9 & 77.8 & 0 \\ 
        acute posthaemorrhagic anaemia & 75 & 57.7 & 0 & 0 & 63.5 & 17.3 & 30.8 \\ 
        gastrooesophageal reflux disease & 74.3 & 28.6 & 0 & 20 & 85.7 & 0 & 22.9 \\ 
        peritonitis & 74.1 & 74.1 & 0 & 29.6 & 96.3 & 51.9 & 70.4 \\ 
        atrial fibrillation & 73.5 & 87.4 & 5.5 & 18.7 & 77.4 & 55.5 & 71 \\ 
        encephalopathy due to toxicity & 73.2 & 7 & 0 & 2.8 & 40.8 & 28.2 & 8.5 \\ 
        sepsis without septic shock & 72.9 & 14.2 & 10.7 & 0 & 47.4 & 17.8 & 20.1 \\ 
        multiple sclerosis & 71.9 & 100 & 75 & 87.5 & 81.3 & 96.9 & 96.9 \\ 
        malignant neoplasm metastasis in brain & 71.1 & 7.9 & 15.8 & 10.5 & 7.9 & 0 & 68.4 \\ 
        primary neoplasms of meninges & 69.4 & 63.9 & 52.8 & 33.3 & 72.2 & 72.2 & 58.3 \\ 
        human immunodeficiency virus disease without mention of tuberculosis or malaria & 69.2 & 36.5 & 0 & 0 & 19.2 & 32.7 & 5.8 \\ 
        orthostatic hypotension & 69.1 & 70.2 & 13.8 & 1.1 & 78.7 & 43.6 & 23.4 \\ 
        hypo-osmolality or hyponatraemia & 69.1 & 16.5 & 0 & 0 & 17.5 & 0 & 4.1 \\ 
        acute pyelonephritis & 65.5 & 69 & 0 & 0 & 72.4 & 37.9 & 65.5 \\ 
        neutropaenia & 61.5 & 89.2 & 0 & 36.9 & 90.8 & 46.2 & 86.2 \\ 
        venous thromboembolism & 59.4 & 56.5 & 47.8 & 34.8 & 43.5 & 33.3 & 71 \\ 
        hypotension, unspecified & 56.8 & 18.9 & 100 & 100 & 29.7 & 83.8 & 0 \\ 
        other specified cardiac arrhythmia & 56.2 & 46.6 & 0 & 0 & 45.2 & 2.7 & 47.9 \\ 
        pulmonary thromboembolism & 56 & 46.6 & 1.6 & 11.9 & 41.5 & 0 & 37.8 \\ 
        urinary tract infection, site not specified & 55.9 & 31.2 & 0 & 0.8 & 45.9 & 19.9 & 0.5 \\ 
        hyperkalaemia & 52.2 & 73.9 & 21.7 & 0 & 69.6 & 69.6 & 69.6 \\ 
        deep bacterial folliculitis or pyogenic abscess of the skin & 49.2 & 37.3 & 0 & 94.9 & 98.3 & 15.3 & 74.6 \\ 
        acute respiratory failure & 47.5 & 90.7 & 0.8 & 61.9 & 86.4 & 83.9 & 93.2 \\ 
        essential hypertension & 39.5 & 81.6 & 0 & 21.1 & 47.4 & 10.5 & 94.7 \\ 
        pleural effusion & 38.6 & 59.1 & 34.1 & 4.5 & 54.5 & 27.3 & 47.7 \\ 
        pneumonia & 37.8 & 96.1 & 68.2 & 92.6 & 89.1 & 74.5 & 93.2 \\ 
        bacterial pneumonia & 35.9 & 84.6 & 7.7 & 48.7 & 38.5 & 100 & 100 \\ 
        congestive heart failure & 30.1 & 80.7 & 0 & 0 & 77.1 & 1.2 & 18.1 \\ 
        bacteraemia & 17.9 & 24.4 & 14.1 & 0 & 15.4 & 6.4 & 5.1 \\ 
        dehydration & 6.1 & 92.1 & 94.7 & 43 & 88.6 & 70.2 & 96.5 \\ 
        chronic kidney disease & 4.6 & 50.4 & 0.4 & 7.3 & 13.5 & 18.8 & 13.5 \\ 
        acute kidney failure & 3.2 & 30.8 & 0.6 & 27.8 & 30 & 26.4 & 64.2 \\ 
\bottomrule
\end{longtable}

\end{document}